\documentclass[journal]{IEEEtran}
\ifCLASSINFOpdf
\else
\fi

\RequirePackage{fix-cm}

\usepackage{graphicx}
\usepackage{floatrow}
\usepackage{epstopdf}
\usepackage{amssymb}
\usepackage{amsmath}
\usepackage{amsthm}
\newtheorem{proposition}{Proposition}
\usepackage{algorithm}
\usepackage{algorithmic}
\usepackage[numbers,sort&compress]{natbib}

\usepackage{booktabs} 
\usepackage{makecell}
\usepackage{floatrow}
\allowdisplaybreaks[4]

\begin{document}
%
\title{Conceptor Learning for Class Activation Mapping}
%
%
%

\author{
Guangwu Qian,
Zhen-Qun Yang,
Xu-Lu Zhang,
Yaowei Wang, ~\IEEEmembership{Member,~IEEE},
Qing Li,~\IEEEmembership{Fellow,~IEEE},
        and~Xiao-Yong Wei,~\IEEEmembership{Senior Member,~IEEE}
\thanks{Guangwu Qian, and Yaowei Wang are with the Peng Cheng Laboratory, Shenzhen, China}
\thanks{Zhen-Qun Yang is with the Department of Biomedical Engineering, Chinese University of Hong Kong, Kowloon, Hong Kong}
\thanks{
Xiao-Yong Wei (e-mail: cswei@scu.edu.cn) and Xu-Lu Zhang are with the College of Computer Science, Sichuan University, Chengdu 610065, China}
\thanks{
Qing Li and Xiao-Yong Wei are with the Department of Computing, Hong Kong Polytechnic University, Kowloon, Hong Kong}
}

%
%

\markboth{IEEE Transactions on Image Processing}%
{Qian \MakeLowercase{\textit{et al.}}: Conceptor-CAM: Conceptor learning for Visual Explanations in Deep Neural Networks}
%



\maketitle

\begin{abstract}
Class Activation Mapping (CAM) has been widely adopted to generate saliency maps which provides visual explanations for deep neural networks (DNNs). The saliency maps are conventionally generated by fusing the channels of the target feature map using a weighted average scheme. It is a weak model for the inter-channel relation, in the sense that it only models the relation among channels in a contrastive way (i.e., channels that play key roles in the prediction are given higher weights for them to stand out in the fusion). The collaborative relation, which makes the channels work together to provide cross reference, has been ignored. Furthermore, the model has neglected the intra-channel relation thoroughly.
In this paper, we address this problem by introducing Conceptor learning into CAM generation. Conceptor leaning has been originally proposed to model the patterns of state changes in recurrent neural networks (RNNs). By relaxing the dependency of Conceptor learning to RNNs, we make Conceptor-CAM not only generalizable to more DNN architectures but also able to learn both the inter- and intra-channel relations for better saliency map generation.
Moreover, we have enabled the use of Boolean operations to combine the positive and pseudo-negative evidences, which has made the CAM inference more robust and comprehensive.
The effectiveness of Conceptor-CAM has been validated with both formal verifications and experiments on the dataset of the largest scale in literature. The experimental results show that Conceptor-CAM is compatible with and can bring significant improvement to all well recognized CAM-based methods, and has outperformed the state-of-the-art methods by $43.14\%\sim 72.79\%$ ($88.39\%\sim 168.15\%$) on ILSVRC2012 in Average Increase (Drop), $15.42\%\sim 42.55\%$ ($47.09\%\sim 372.09\%$) on VOC, and $17.43\%\sim 31.32\%$ ($47.54\%\sim 206.45\%$) on COCO, respectively.
\end{abstract}


%
\IEEEpeerreviewmaketitle

\section{Introduction}
%
%
%
%
\label{sec:intro}
\IEEEPARstart{V}{isualization} approaches based on Class Activation Mapping (CAM) \cite{7780688} have been widely employed to provide intuitive explanations to DNNs.  CAM-based approaches visualize the importance of neurons of a certain DNN layer (usually the last layer of the feature maps) to the final decisions by generating a saliency map according to the degrees of activation on these neurons regarding the weights, gradients, and/or their changes upon the DNN inferences. Representative approaches include Grad-CAM \cite{selvaraju2017grad} which generates the saliency map  by calculating the weighted average of a feature map over its channels in which the weight of each channel has been obtained by average pooling the gradients with respect to the final decision. Another popular approach is the Score-CAM \cite{wang2020score} which obtains the channel weighs using the global contribution of the corresponding input features instead of the locality sensitive measurements (e.g., the gradients).

Despite the success of CAM-based approaches, we argue that the way of generating the saliency maps by fusing the activation evidences (e.g., gradients in Grad-CAM) over channels using the average pooling has neglected the nature of using ``cross-evidences'' for inference in DNNs. That is, channels of a DNN are usually trained to respond to various types of evidences \cite{zeiler2014visualizing}(i.e., intra-channel patterns such as the skin colors, face shapes, or eyelids in a DNN for facial recognition tasks), on the basis of which the CNN model then
learns the inter-channel patterns by assembling these evidences across channels synthetically. Therefore, the weighted average fusion used in conventional CAM-based approaches is not sophisticated enough to model the inter-channel assembling. Furthermore, the single-valued average of all the gradients of a channel is far from adequate to model the intra-channel patterns (i.e., the relations among neurons of each channel).

Besides investigating the evidences from a channel perspective, recent studies \cite{wei2017object,8237644,li2018tell} have found that the CAM results can be improved by distinguishing the target (or positive) evidences from the non-target (or negative) evidences. For example, Kim et al. has proposed a two-stage learning method in \cite{li2018tell} which in the first stage, the CNN model has been trained to generate a saliency map with Grad-CAM and the map will be used as an indication of positive evidences for the inference (e.g., for recognizing a ``car'' object). In the second stage, the negative map will be generated by inverting the positive map and used as a mask applied to the original image, with which the network will be retrained with an inverted label (e.g., for recognizing the ``non-car'' region). The experimental results show that the final positive saliency map can focus more on the target region than that of conventional CAM-based methods. However, the use of positive and negative evidences is based on heuristics, which is straightforward but not explicitly formulated.

\begin{figure*}[t]
	\centering
	\includegraphics[width=0.96\textwidth]{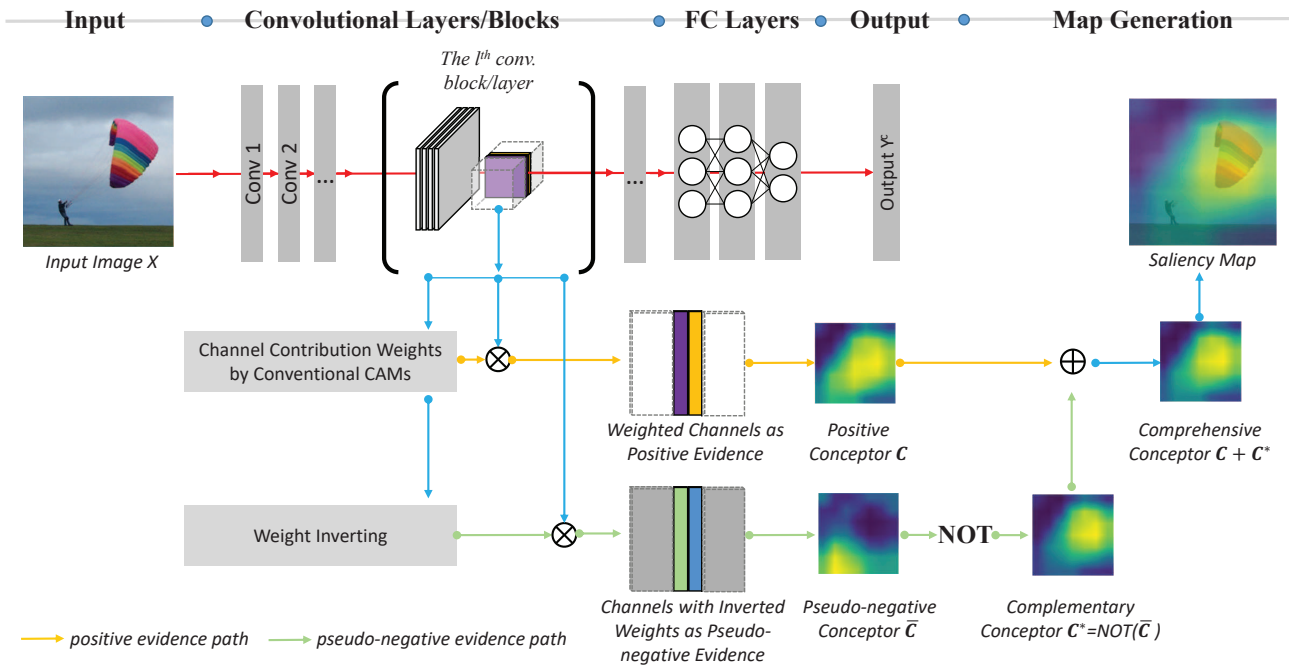}
	\caption{Conceptor Learning for Class  Activation  Mapping (CAM). Feature map at the $l^{th}$ block/layer is first fed into the conventional 
	CAM methods to evaluate the channel contribution, with which the channel maps are weighted and vectorized as the evidences for Conceptor learning. This results in a positive Conceptor $\boldsymbol{C}$ which has been encapsulated with the inter- and intra-channel relations. In a similar way, the pseudo-negative evidences can be generated for learning the pseudo-negative Concepotr $\bar{\boldsymbol{C}}$ which will be inverted using the NOT operation 
to create a complementary Conceptor that has interpreted the channel relation from a different point of view. The positive and complementary Conceptors are finally fused to generate a saliency map which provides a more comprehensive visual explanation of the DNN inference. Note in the parachute example, the Conceptor learning can fix the false attention on the part between the person and the parachute (a typical result of the conventional CAM-based methods).}
\label{fig:conceptor-CAM-arch}
\end{figure*}

In this paper, we propose to address the aforementioned issues within a unified framework and in a formulated way by adopting the \textit{Conceptor} learning \cite{jaeger2014controlling,jaeger2017using}. The \textit{Conceptor} learning is a method that has been originally proposed by Jaeger \cite{jaeger2014controlling} to learn the pattern of how the neuron states of a RNN model change over time, and the learned pattern (encapsulated in a Conceptor matrix $\boldsymbol{C}$) can then be used to reproduce the future neuron states even with missing or incomplete inputs. 
As shown in Fig.~\ref{fig:conceptor-CAM-arch}, we will generate Conceptors to model the inter- and intra- channel relations and use them to synchronize the channel vectors. The Conceptors can be learned from the positive and pseudo-negative evidences separately and then be fused to create more comprehensive saliency maps.
By adopting the \textit{Conceptors} into CAM learning, it brings forth several advantages as follows.
\begin{itemize}
    \item \textbf{Inter-channel pattern modeling}: We have released the dependence of Conceptor learning on RNN models, so that we can adopt the way for modeling the pattern among neuron states to model that of the channel states. This is straightforward when the channel states are converted into state vectors (see Section~\ref{sec:regulation}).
    \item \textbf{Intra-channel pattern modeling}: We will provide a formal verification in Section~\ref{sec:intra-channel} to show that the Conceptor matrix $\boldsymbol{C}$ indeed also models the patterns among the elements of a state vector, and therefore, using $\boldsymbol{C}$ to refine the state vectors will be more robust to noise because the new vectors have been generated with the reference to both inter- and intra-channel relations.
    \item \textbf{Boolean reasoning}: It has been justified in \cite{jaeger2017using} that Boolean operations are well defined on Conceptors. It is thus more formulated and convenient to calculate or combine different evidences (e.g., positive or negative, target or non-target) by using Boolean operations. We will show in Section~\ref{sec:comprehensive_conceptor-cam} that Conceptor learning can be used to combine positive and pseudo-negative evidences for a more comprehensive modeling of the CAM inference.  
\end{itemize}

\section{Related Work}
\label{sec:related_work}
The proposed framework is related to two groups of methods, namely the CAM-based methods \cite{7780688,selvaraju2017grad,wang2020score} and the Conceptor learning \cite{jaeger2017using}, in the way that the proposed method aims to improve the performance of CAM-based methods with the Conceptor learning as the backbone.
\subsection{CAM-based Methods}
\label{sec:cam}
The original CAM has been proposed by Zhou et al. \cite{7780688} for identifying the salient regions of an image that have supported the prediction. The assumption is that the neurons, which have been activated more sufficiently than others, will contribute more for the prediction. The evaluation of activation can first be done at the last convolutional layer of a CNN model through the weights assigned onto the connections of the target convolutional neurons to these at the first fully connected (FC) layer. The contribution of a convolutional channel can then be summarized by taking the average of activation values (the connection weights) of its member neurons, with which the feature maps of channels are fused into the spatial dimensions. Finally, the fused feature map will be re-scaled into the same (spatial) size of the input image as a saliency map indicating the contributions of image regions. As the first CAM method, the original CAM has provided a new tool to investigate how deep neural networks conduct the inference, and thus has attracted intensive research attention. However, the original CAM is less feasible than its succeeders, because its way of weight summarization requires to physically insert an average pooling layer between the last convolutional layer and the first FC layer which will introduce the complication of network architecture modification.

Grad-CAM proposed by Selvaraju et al. \cite{selvaraju2017grad} is the best known succeeder of the original CAM. It has eliminated the requirement of architecture modification by calculating the channel weights from the gradients directly (instead of the connection weights used by the original CAM). This makes it compatible to most of the CNN models, and thus it has been widely adopted.
However, the dependence on the gradients may also introduce the saturation and confidence problems \cite{wang2020score} because of the noise and vanishing issues.
Score-CAM \cite{wang2020score} is proposed by Wang et al. to get rid of the dependence using the perturbation which evaluates the channel contribution by observing the change of the responses (i.e., feature maps) by adding a small ``disruption'' to the input (perturbing). More specifically, the feature map of a channel will be used as a mask on the input image, and the masked image will be feed into the networks to obtain a feature map on this channel. The contribution weight of the channel is then evaluated by the change on the feature map it has made by masking the input.
In addition to these CAM-based methods, a lot of other variations have been proposed recently such as Smooth Grad-CAM++ \cite{8354201}, SS-CAM \cite{wang2020ss}, Layer-CAM \cite{9462463}, and Ablation-CAM \cite{desai_2020_WACV}. 
Besides, there are many other CAM-based methods in which the saliency maps are generated for specific tasks (e.g., label-free localization \cite{9385115}, high-quality proposal \cite{9069411}, visualization of deep reinforcement learning \cite{8847950}, localization comparison \cite{8821313}, scaling method comparison \cite{pmlr-v97-tan19a}, and performance evaluation by experts \cite{ozturk2020automated,GARGEYA2017962}).
We will skip the details due to space limitation.

In this paper, we aim to build a synchronizer on top of CAM-based methods, which can synchronize the channel vectors for better saliency map generation by leveraging the inter- and intra-channel relations. The synchronization can be conducted in a formulated way with the support of Conceptor learning \cite{jaeger2017using}.
In terms of inter-channel relation, the Conceptor learning has introduced the collaborative modeling in addition to the contrastive modeling used by most CAM-based methods.
In terms of intra-channel relation, the Conceptor learning can also model the collaborative relation of ``pixels'' inside a channel. We will provide a formal verification in Section~\ref{sec:intra-channel}.
In fact, the authors of Grad-CAM++ \cite{8354201}, which is considered as an upgraded version of the famous Grad-CAM, have noticed that calculating the contribution weight by simply using the average may cause the small-but-important regions to be disregarded, and thus have proposed a better scheme to give credits to such regions for improving the performance. We consider this work as towards the same direction of better intra-channel relation modeling. 

\subsection{Conceptor Learning}
\label{sec:conceptor}
Conceptor learning \cite{jaeger2017using} is a technique to model the patterns of neuron state changes with a matrix called Conceptor \cite{jaeger2017using}. The learned Conceptor matrix can be used to either regulate the original state vectors or predict the new state.
By nature, it is more about a way to improve the performance of existing networks from inside rather than an independent method for specific tasks.
For example, it has been employed for improving the classification performance of RNNs \cite{WANG2016237} and CNNs \cite{QIAN20181034,10.1007/978-3-319-59081-3_35}. It has also been applied to the time series prediction \cite{zhang2020chaotic,8894427}, to overcome the catastrophic interference in multi-task learning \cite{he2018overcoming}, and to improve the performance of cache-based communications \cite{7875131}.

Conceptor learning is related to the proposed method straightforwardly, because we have employed it for channel (state) relation modeling. Another advantage is that the Boolean operators (AND, OR, NOT) have been well defined on Conceptors, which means we can easily find the positive, negative, or joint evidences that has supported the prediction by using these operators. This gives us the chance to investigate the networks from different perspectives.
Exemplar applications of this property include the Conceptor-based post-processing of word vectors in \cite{Liu_Ungar_Sedoc_2019} and the multi-label classification in \cite{qian2019single}.
In Conceptor-CAM, we will use it for converting the pseudo-negative evidences to pseudo-positive evidences using NOT operation. More details will be given in Section~\ref{sec:pseudo-negative}.

\section{Conceptor-based Class Activation Mapping}
\label{sec:conceptor-cam}
\subsection{Formulation for CAM-based Methods}
\label{sec:formulation}
To facilitate the description, let us formulate conventional CAM-based methods into a unified framework before we go into more details.
Denote a CNN model as a function $Y^c=f(X):\mathbb{R}^d\to \mathbb{R}$
which maps a $d$-dimensional input $X\in\mathbb{R}^d$ into a probability $Y^c\in \mathbb{R}$ for the class $c$,
CAM-based methods are indeed following the same principle to generate a saliency map $\boldsymbol{\mathcal{S}}_l^c$ from an activation map $\boldsymbol{\mathcal{F}}_l^c$ (i.e., the feature map) at the $l^{th}$ layer of the convolution stage by a weighted average over all $\boldsymbol{\mathcal{F}}_l^c$'s channels as
\begin{equation}
\label{eqn:general_cam}
    \boldsymbol{\mathcal{S}}_l^c=\mathbf{\Psi}(\boldsymbol{\mathcal{F}}_l^c\boldsymbol{\vec{w}}_l^c)
\end{equation}
where $\boldsymbol{\vec{w}_l^c}$ is the weight vector for channels, and $\mathbf{\Psi}(\cdot)$ is a function that re-scales the input into the same (spatial) size as the input $X$. The re-scaling usually includes a set of operations such as ReLU, reshaping, up-sampling, and normalization.
Note that, to facilitate the illustration, we have reshaped each channel as a column vector instead of using its matrix form in previous studies. Therefore, $\boldsymbol{\mathcal{F}}_l^c\in\mathbb{R}^{M\times K}$  is a matrix consisting of these channel vectors ($M$ and $K$ are the length and number of channel vectors respectively).

With Eq.~(\ref{eqn:general_cam}), it is easier to see that the CAM-based methods are different from each other regrading how the weight vector $\boldsymbol{\vec{w}}_l^c$ has been generated. Let us add the algorithm indicators $C$, $G$, $S$ for CAM, Grad-CAM, and Score-CAM respectively, and the CAM-based methods can be rewritten as follows
\begin{equation}
\label{eqn:cam-based}
    \begin{split}
    \text{CAM}:\hspace{0.1in}\boldsymbol{\vec{w}}_l^c|C=&\frac{1}{N_{l+1} N_{l+2}}{\left(\boldsymbol{W}^c_{l+1}\right)}^\top\boldsymbol{\vec{1}},\\
    &\hspace{0.1cm}\boldsymbol{W}^c_{l+1}\in \mathbb{R}^{N_{l+2}\times N_{l+1}},\\
    \text{Grad-CAM}:\hspace{0.1in}\boldsymbol{\vec{w}}_l^c|G=&\frac{1}{KM}\left(\frac{\partial Y^c}{\partial \boldsymbol{\mathcal{F}}_{l}^c}\right)^\top\boldsymbol{\vec{1}},\hspace{0.1cm}\boldsymbol{\mathcal{F}}_l^c\in\mathbb{R}^{M\times K},\\
    \text{Score-CAM}:\hspace{0.1in}\boldsymbol{\vec{w}}_l^c|S=&(w_k)=(f(X\circ \boldsymbol{\mathcal{H}}_{l\cdot k}^c)-f(X)),\\
    \hspace{0.1in}\boldsymbol{\mathcal{H}}_{l\cdot k}^c=&\mathbf{\Psi}(\boldsymbol{\mathcal{F}}_{l\cdot k}^c), \hspace{0.1cm}k\in [1,K]
    \end{split}
\end{equation}
where $\boldsymbol{\vec{1}}$ denotes all-ones column vectors, $\top$ denotes the matrix transpose operator, $\boldsymbol{W}^c_{l+1}$ is the weight matrix encapsulated with the weights that connect the $N_{l+1}$ neurons at the last convolution layer to the $N_{l+2}$ neurons at the first full-connected layer, $k$ is an iterator for traversing channels, and $\circ$ denotes the Hadamard product.

\subsection{Channel Synchronization with Conceptor Learning}
\label{sec:regulation}
With Eq.~(\ref{eqn:cam-based}), we can see that different CAM-based methods are indeed measuring the contributions of channels to the inference by collecting evidences from different sources (e.g., the weights, the gradients, the feature map itself). These sources may have their own advantages over one another, but they share a common nature of providing a ``contrastive'' evaluation of the inter-channel relation rather than modeling the ``collaborative'' relation. When the channels are fused in Eq.~(\ref{eqn:general_cam}), the opportunity of modeling this collaborative relation is gone. Therefore, we propose to model the relation before the fusion. It can be written as
\begin{equation}
\label{eqn:conceptor-cam}
    \boldsymbol{\mathcal{S}}_l^{c*}=\mathbf{\Psi}\left(\underbrace{\vphantom{\Big[]}\boldsymbol{{C}}\overbrace{\vphantom{\Big[]}\left[\boldsymbol{\mathcal{F}}_l^c diag(\boldsymbol{\vec{w}}_l^c)\right]}^{\text{weighting only}}}_{synchronization}\hspace{0.1cm}\boldsymbol{\vec{1}}\right),
\end{equation}
where $diag(\cdot)$ is a standard function to convert a vector into a diagonal matrix, and $\boldsymbol{{C}}$ is a matrix which has been encapsulated with the inter-channel relation and thus has been used to synchronize the (weighted) channels.
The difference of Eq.~(\ref{eqn:conceptor-cam}) to Eq.~(\ref{eqn:general_cam}) is that we have weighted the channels with weights in $\boldsymbol{\vec{w}}_l^c$ (can be learned from any existing CAM-based method) to make the channels that include the key evidences ``stand out'' from other channels, but the weighted channels have not been fused directly as in Eq.~(\ref{eqn:general_cam}). The fusion will be conducted after the ``synchronization'' with respect to the inter- and intra-channel relations encapsulated in the Conceptor matrix $\boldsymbol{{C}}\in\mathbb{R}^{M\times M}$.

\subsubsection{Learning Inter-Channel Relation with Conceptors}
\label{sec:inter-channel}
It is straightforward to adopt the way of learning the inter-neuron relation in Conceptor learning to our goal of modeling inter-channel relation. More specifically, Conceptor learning is to find a Concept matrix $\boldsymbol{{C}}\in\mathbb{R}^{M\times M}$ which is a transformation matrix to synchronize a set of neuron state vectors $\boldsymbol{Z}\in\mathbb{R}^{M\times K}$ (in our case $\boldsymbol{Z}=\boldsymbol{\mathcal{F}}_l^c diag(\boldsymbol{\vec{w}}_l^c)\in\mathbb{R}^{M\times K}$) so that the loss of the synchronization can be minimized. The loss function is written as
\begin{equation}
\label{eqn:loss}
\mathcal{L} = \mathbb{E}\left[\big\Vert \boldsymbol{{Z}}-\boldsymbol{{C}}\boldsymbol{Z} \big\Vert_{fro}^2\right] + \alpha^{-2}\|\boldsymbol{{C}}\|_{fro}^2,
\end{equation}
where the $\|\cdot\|^2_{fro}$ denotes the Frobenius norm, and $\alpha\in(0,\infty)$ is a balancing factor between the direct loss of the transformation (the first part of Eq.~(\ref{eqn:loss})) and the parameter regulator (the second part). Note that the $\|\boldsymbol{{C}}\|_{fro}^2$ in fact serves to prevent the transformation matrix $\boldsymbol{{C}}$ from falling into an identity matrix. Otherwise, there is no synchronization taken place (i.e., $\boldsymbol{{Z}}=\boldsymbol{{C}}\boldsymbol{Z}$).

The rationale behind Eq.~(\ref{eqn:loss}) is to ``synchronize'' the state vectors with the help of their relation to other state vectors (represented by the Conceptor matrix $\boldsymbol{{C}}$). Therefore, the synchronized vectors, which have been generated with the reference to other vectors, are expected to be more reliable than the original vectors (which are subject to the random noise).
Moreover, the claim that $\boldsymbol{{C}}$ is a representation of the inter-vector relation is self-explanatory from the solution
 \begin{equation}
\label{eqn:con}
\boldsymbol{{C}} = \boldsymbol{{R}}(\boldsymbol{{R}}+\alpha^{-2}\boldsymbol{I})^{-1},
\end{equation}
where $\boldsymbol{R}=\mathbb{E}[\boldsymbol{{Z}}(\boldsymbol{{Z}})^\top]\in\mathbb{R}^{M\times M}$ is a (inter-channel) correlation matrix and $\boldsymbol{I}\in\mathbb{R}^{M\times M}$ is an identity matrix.
%
%
It has been proved in \cite{jaeger2017using} that Eq.~(\ref{eqn:con}) is a unique and optimal solution to minimize the objective function Eq.~(\ref{eqn:loss}). We will skip the details due to space limitation. The formal verification can be found in \cite{jaeger2017using}.

\subsubsection{$\boldsymbol{{C}Z}$ as an Intra-channel Co-Reconstruction}
\label{sec:intra-channel}
After the Conceptor matrix $\boldsymbol{{C}}$ has been learned with the inter-channel relation (encapsulated in $\boldsymbol{R}$), it is easy to see that the transformation $\boldsymbol{{C}Z}$ equals to an intra-channel reconstruction process which has been implemented by generating each element of the new channel vector with the weighted average of other elements from the old channel vector (i.e., $(\boldsymbol{{C}Z})_{ij}=\boldsymbol{C}_{i:} \boldsymbol{Z}_{:j}$ where $:$ denotes a row/column
collector).
Beyond the intuition of the calculation, let us verify this formally because it is not a claim of the original Conceptor learning in \cite{jaeger2017using}.

Let $\boldsymbol{Z}_{i:=0}\in\mathbb{R}^{M\times K}$ denote a matrix generated by setting the $i^{th}$ row of $\boldsymbol{Z}$ to zeros, and $\boldsymbol{Z}_{\bar{i:}=0}\in\mathbb{R}^{M\times K}$ denote a complementary matrix of $\boldsymbol{Z}_{i:=0}$ generated by setting all rows of $\boldsymbol{Z}$ to zeros except the $i^{th}$.
Assume there is a matrix $\hat{\boldsymbol{C}}\in\mathbb{R}^{M\times M}$ that is able to minimize the intra-channel reconstruction cost of recovering $\boldsymbol{Z}$ from $\boldsymbol{Z}_{i:=0}$ (i.e., recovering the $i^{th}$ elements of all channel vectors) through the transformation $\hat{\boldsymbol{C}}\boldsymbol{Z}_{i:=0}$, the cost function can be written as
\begin{equation}
\label{eqn:loss_*}
\hat{\mathcal{L}} = \mathbb{E}\left[\left\Vert \frac{1}{M-1}\sum_{i=1}^{M}(\boldsymbol{Z}-\hat{\boldsymbol{C}}\boldsymbol{Z}_{i:=0} )\right\Vert_{fro}^2\right] + \alpha^{-2}\|\hat{\boldsymbol{C}}\|_{fro}^2,
\end{equation}
where the first part is an unbiased estimation of the reconstruction cost, and the second part is the parameter regulator. We will prove that
\begin{proposition}
The Conceptor matrix $\boldsymbol{C}$ learned through the inter-channel relation is also a solution to minimize the intra-channel reconstruction cost $\hat{\mathcal{L}}$.
%
\end{proposition}
\begin{proof}
Let us expand $\hat{\mathcal{L}}$ for easier calculation first as
\begin{align*}
\label{eqn:prop}
\hat{\mathcal{L}} =& \mathbb{E}\left[\left\Vert \frac{1}{M-1}\sum_{i=1}^{M}\Big(\boldsymbol{Z}-\hat{\boldsymbol{C}}\boldsymbol{Z}_{i:=0} \Big)\right\Vert_{fro}^2\right] + \alpha^{-2}\|\hat{\boldsymbol{C}}\|_{fro}^2\\
=& \mathbb{E}\left[\left\Vert \frac{1}{M-1}\left(\sum_{i=1}^{M}\boldsymbol{Z}-\sum_{i=1}^{M}\hat{\boldsymbol{C}}\boldsymbol{Z}_{i:=0}\right)\right\Vert_{fro}^2\right] +\\& \alpha^{-2}\|\hat{\boldsymbol{C}}\|_{fro}^2\\
=& \mathbb{E}\left[\left\Vert \frac{1}{M-1}\left(M\boldsymbol{Z}-\sum_{i=1}^{M}\hat{\boldsymbol{C}}\Big(\boldsymbol{Z}-\boldsymbol{Z}_{\bar{i:}=0}\Big)\right)\right\Vert_{fro}^2\right] +\\& \alpha^{-2}\|\hat{\boldsymbol{C}}\|_{fro}^2\\
=& \mathbb{E}\left[\left\Vert \frac{1}{M-1}\left(M\boldsymbol{Z}-M\hat{\boldsymbol{C}}\boldsymbol{Z}+\sum_{i=1}^{M}\Big(\hat{\boldsymbol{C}}\boldsymbol{Z}_{\bar{i:}=0}\Big)\right)\right\Vert_{fro}^2\right] \\&+ \alpha^{-2}\|\hat{\boldsymbol{C}}\|_{fro}^2\\
=& \mathbb{E}\left[\left\Vert \frac{1}{M-1}\left(M\boldsymbol{Z}-M\hat{\boldsymbol{C}}\boldsymbol{Z}+\hat{\boldsymbol{C}}\boldsymbol{Z}\right)\right\Vert_{fro}^2\right] +\\& \alpha^{-2}\|\hat{\boldsymbol{C}}\|_{fro}^2\\
=& \mathbb{E}\left[\left\Vert \frac{M}{M-1}\boldsymbol{Z}-\hat{\boldsymbol{C}}\boldsymbol{Z}\right\Vert_{fro}^2\right] + \alpha^{-2}\|\hat{\boldsymbol{C}}\|_{fro}^2\\
=& \mathbb{E}\left[\left\Vert \lambda\boldsymbol{Z}-\hat{\boldsymbol{C}}\boldsymbol{Z}\right\Vert_{fro}^2\right] + \alpha^{-2}\|\hat{\boldsymbol{C}}\|_{fro}^2\hspace{0.1in} \\& \text{where }\lambda=\frac{M}{M-1}\\
=& tr\Big(\big((\lambda\boldsymbol{I}-\hat{\boldsymbol{C}})\boldsymbol{Z}\big)^\top\big((\lambda\boldsymbol{I}-\hat{\boldsymbol{C}})\boldsymbol{Z}\big)\Big) +\\& tr\Big(\alpha^{-2}(\hat{\boldsymbol{C}})^\top\hat{\boldsymbol{C}}\Big)\\
=& tr\Big(\big(\lambda\boldsymbol{I}-(\hat{\boldsymbol{C}})^\top\big)\big(\lambda\boldsymbol{I}-\hat{\boldsymbol{C}}\big)\boldsymbol{R} +\\& \alpha^{-2}(\hat{\boldsymbol{C}})^\top\hat{\boldsymbol{C}}\Big)\\
=& tr\big(\lambda^2\boldsymbol{R}-\lambda(\hat{\boldsymbol{C}})^\top\boldsymbol{R}-\lambda\hat{\boldsymbol{C}}\boldsymbol{R} +\\& (\hat{\boldsymbol{C}})^\top\hat{\boldsymbol{C}}(\boldsymbol{R}+ \alpha^{-2}\boldsymbol{I})\big)\\
=& \sum_{i=1}^{M}\boldsymbol{e}_i^\top\big(\lambda^2\boldsymbol{R}-\lambda(\hat{\boldsymbol{C}})^\top\boldsymbol{R}-\lambda\hat{\boldsymbol{C}}\boldsymbol{R} +\\& (\hat{\boldsymbol{C}})^\top\hat{\boldsymbol{C}}(\boldsymbol{R}+ \alpha^{-2}\boldsymbol{I})\big)\boldsymbol{e}_i.\\
\end{align*}
Let $\boldsymbol{A}=\boldsymbol{R}+ \alpha^{-2}\boldsymbol{I}$, for the $kl^{th}$ element of  $\hat{\boldsymbol{C}}$, we have
\begin{equation}
\label{eqn:gradient}
\begin{split}
\frac{\partial \hat{\mathcal{L}}}{\partial \hat{\boldsymbol{C}_{kl}}}=&-2\lambda\boldsymbol{R}_{kl}+\sum_{i=1}^{M}\left(\frac{\partial \sum_{j,a=1,\cdots,M}\hat{\boldsymbol{C}_{ij}}\hat{\boldsymbol{C}_{ja}}\boldsymbol{A}_{ai}}{\partial \hat{\boldsymbol{C}_{kl}}}\right)\\
=&-2\lambda\boldsymbol{R}_{kl}+ \sum_{i=1}^{M}\left(\sum_{a=1}^{M}\frac{\partial \hat{\boldsymbol{C}_{ki}}\hat{\boldsymbol{C}_{ka}}\boldsymbol{A}_{ai}}{\partial \hat{\boldsymbol{C}_{kl}}}\right)\\
=&-2\lambda\boldsymbol{R}_{kl}+ \sum_{a=1}^{M}( \hat{\boldsymbol{C}_{ka}}\boldsymbol{A}_{al})+\sum_{i=1}^{M}(\hat{\boldsymbol{C}_{ki}}\boldsymbol{A}_{li})\\
=&-2\lambda\boldsymbol{R}_{kl}+ ( \hat{\boldsymbol{C}}\boldsymbol{A})_{kl}+(\hat{\boldsymbol{C}}\boldsymbol{A})_{kl}\\
=&-2\lambda \boldsymbol{R}_{kl}+2(\hat{\boldsymbol{C}}\boldsymbol{A})_{kl}.
\end{split}
\end{equation}
Therefore, by collecting all elements into a matrix form, we have 
$\frac{\partial \hat{\mathcal{L}}}{\partial \hat{\boldsymbol{C}}}=-2\lambda \boldsymbol{R}+2(\hat{\boldsymbol{C}}\boldsymbol{A})$,
and the solution can then be found at $\frac{\partial \hat{\mathcal{L}}}{\partial \hat{\boldsymbol{C}}}=0$ as
\begin{equation}
\label{eqn:con_*}
    \hat{\boldsymbol{C}}=\lambda\boldsymbol{{R}}(\boldsymbol{{R}}+\alpha^{-2}\boldsymbol{I})^{-1}=\lambda\boldsymbol{C} \hspace{0.1in}\text{ where } \lim_{M\to\infty}\lambda=1.
\end{equation}
It is easy to see that $\boldsymbol{C}$ is also a solution to $\hat{\mathcal{L}}$.
\end{proof}

\subsection{Collection of Pseudo-Negative Evidence}
\label{sec:pseudo-negative}
Negative evidences that lead to the inverse prediction of class-$c$ (i.e., non-$c$, denote as $\bar{c}$ hereafter) are often useful to provide a cross-reference for the prediction. Due to that the collection of such evidences is infeasible at feature map layers, we propose to collect and build pseudo-evidences in this section.
The idea is to reverse the weights learned by CAM-based methods (i.e., Eq.~(\ref{eqn:cam-based}) so that the key channels supporting the prediction of $c$ will be suppressed while those less important channels will be given with key roles in the fusion (i.e., Eq.~(\ref{eqn:general_cam})). A pseudo-negative Concept matrix $\bar{\boldsymbol{C}}$ can then be learned to modeling the evidences that is not predicting class $c$.
This process can be formulated as
\begin{equation}
\label{eqn:pseudo-negative}
\begin{split}
\bar{\boldsymbol{{C}}} =& \bar{\boldsymbol{{R}}}(\bar{\boldsymbol{{R}}}+\alpha^{-2}\boldsymbol{I})^{-1},\\
\bar{\boldsymbol{R}}=&\mathbb{E}[\bar{\boldsymbol{Z}}(\bar{\boldsymbol{Z})}^\top]\in\mathbb{R}^{M\times M},\\
\bar{\boldsymbol{Z}}=&\boldsymbol{\mathcal{F}}_l^c diag(\boldsymbol{\vec{1}}-\boldsymbol{\vec{w}}_l^c)\in\mathbb{R}^{M\times K}
\end{split}
\end{equation}
where the weight reversing has been implemented by $\boldsymbol{\vec{1}}-\boldsymbol{\vec{w}}_l^c$ which inverts the influences of channels to prediction as well.

\subsection{Comprehensive Conceptor-CAM}
\label{sec:comprehensive_conceptor-cam}
Eq.~(\ref{eqn:pseudo-negative}) for obtaining the pseudo-negative Conceptor $\bar{\boldsymbol{C}}$ in fact has provided another viewpoint of looking at the evidences stored in the channels. Therefore, we can reverse the Conceptor to obtain a positive evidence collected from a different point of view and use it as a complementary source to generate the saliency map. This can be easily implemented with the Concepotors on which Boolean operations (AND, OR, NOT) are well defined. Denote the complementary Conceptor as $\boldsymbol{C}^*$, we have
\begin{equation}
\label{eqn:con_com}
    \boldsymbol{C}^*=\neg(\bar{\boldsymbol{C}})=\bar{\boldsymbol{R}}^{-1}(\bar{\boldsymbol{{R}}}^{-1}+\alpha^{2}\boldsymbol{I})^{-1},
\end{equation}
where $\neg(\cdot)$ denotes the NOT operator defined on Conceptors \cite{jaeger2017using}.

Finally, we can generate a more comprehensive saliency map with $\boldsymbol{C}$ and $\boldsymbol{C}^*$ for positive evidences of the prediction as
\begin{equation}
\label{eqn:con_fused}
    \boldsymbol{\mathcal{S}}_l^{c+}=\mathbf{\Psi}\left(\underbrace{\vphantom{\Big[]}\frac{1}{2}(\boldsymbol{{C}}+\boldsymbol{C}^*)}_{fusion}\boldsymbol{Z}\hspace{0.1cm}\boldsymbol{\vec{1}}\right).
\end{equation}
Putting all together, Algorithm~\ref{algo:conceptor-cam} has depicted the process of calculating the comprehensive Conceptor-CAM.

\begin{algorithm}[!htb]
\caption{Conceptor-CAM algorithm}
\label{algo:conceptor-cam}
\begin{algorithmic}[1]
\REQUIRE ~~\\
Image $X$, Model $f(x)$, class $c$, layer $l$
\ENSURE ~~\\
The saliency map $\boldsymbol{\mathcal{S}}_l^{c+}$
\STATE $\boldsymbol{\mathcal{F}}_l \leftarrow f_l(X)$ //Obtain the feature map at the layer $l$
\STATE $K \leftarrow$ the number of channels in $\boldsymbol{\mathcal{F}}_l$
\FOR {$k \in [1,K]$}
    \STATE $\boldsymbol{\mathcal{H}}_{l \cdot k} \leftarrow Rescale(\boldsymbol{\mathcal{F}}_{l \cdot k})$
    \STATE $\boldsymbol{\vec{w}}_k \leftarrow f^c(X\circ \boldsymbol{\mathcal{H}}_{l \cdot k})$ // $f^c(\cdot)$ is the largest score corresponding to class $c$
\ENDFOR
\STATE $\boldsymbol{Z} \leftarrow \boldsymbol{\mathcal{F}}_l diag(\boldsymbol{\vec{w}}_l^c), \bar{\boldsymbol{Z}} \leftarrow \boldsymbol{\mathcal{F}}_l diag(\boldsymbol{\vec{1}}-\boldsymbol{\vec{w}}_l^c)$
\STATE $\boldsymbol{R} \leftarrow \mathbb{E}[\boldsymbol{{Z}}(\boldsymbol{{Z}})^\top], \bar{\boldsymbol{R}} \leftarrow \mathbb{E}[\bar{\boldsymbol{Z}}(\bar{\boldsymbol{Z})}^\top]$
\STATE $\boldsymbol{{C}} \leftarrow \boldsymbol{{R}}(\boldsymbol{{R}}+\alpha^{-2}\boldsymbol{I})^{-1}, \bar{\boldsymbol{{C}}} \leftarrow \bar{\boldsymbol{{R}}}(\bar{\boldsymbol{{R}}}+\alpha^{-2}\boldsymbol{I})^{-1}, \boldsymbol{C}^* \leftarrow \neg(\bar{\boldsymbol{C}}) \leftarrow \bar{\boldsymbol{R}}^{-1}(\bar{\boldsymbol{{R}}}^{-1}+\alpha^{2}\boldsymbol{I})^{-1}$
\STATE $\boldsymbol{\mathcal{S}}_l^{c+} \leftarrow \mathbf{\Psi}\left(\frac{1}{2}{\vphantom{\Big[]}(\boldsymbol{{C}}+\boldsymbol{C}^*)}\boldsymbol{Z}\hspace{0.1cm}\boldsymbol{\vec{1}}\right)$
\end{algorithmic}
\end{algorithm}

\begin{table*}[htbp]
\caption{Test of Compatibility of Conceptor Learning to Conventional CAM-based Methods. The best results are in bold font.}
\label{tbl:cam-based_methods}
\begin{center}
\begin{small}
\resizebox{\linewidth}{!}{
\begin{tabular}{c|c|c|c|c|c|c|c|c|c|c|c|c}
\hline
& \multicolumn{2}{c|}{Grad-CAM} & \multicolumn{2}{c|}{Grad-CAM++} & \multicolumn{2}{c|}{Score-CAM} &  \multicolumn{2}{c|}{Layer-CAM} & \multicolumn{2}{c|}{SS-CAM} & \multicolumn{2}{c}{Ablation-CAM}\\
\cline{2-13}
  Runs & \makecell[c]{Average\\Increase $\uparrow$} & \makecell[c]{Average\\Drop $\downarrow$} & \makecell[c]{Average\\Increase $\uparrow$} & \makecell[c]{Average\\Drop $\downarrow$} & \makecell[c]{Average\\Increase $\uparrow$} & \makecell[c]{Average\\Drop $\downarrow$} & \makecell[c]{Average\\Increase $\uparrow$} & \makecell[c]{Average\\Drop $\downarrow$} & \makecell[c]{Average\\Increase $\uparrow$} & \makecell[c]{Average\\Drop $\downarrow$} & \makecell[c]{Average\\Increase $\uparrow$} & \makecell[c]{Average\\Drop $\downarrow$} \\
\hline
\makecell[c]{without\\Conceptors} & 33.45 & 19.24 & 31.68 & 19.84 & 38.11 & 13.96 & 31.57 & 19.87 & 32.99 & 18.42 & 32.43 & 19.48\\
\hline
\makecell[c]{with\\Conceptors} &  31.10 & 20.63 & 31.29 & 20.47 & 40.93 & 11.51 & 31.37 & 20.49 & 34.95 & 16.61 & 30.57 & 20.91\\
\hline
\makecell[c]{with Tanh+\\Conceptors} &  44.22 & 9.76 & 44.94 & 9.82 & \textbf{51.39} & \textbf{6.70} & 45.66 & 9.28 & 46.18 & 8.52 & 46.09 & 7.84\\
\hline
\makecell[c]{Improvement\\($\%$)} & 32.20 & 97.13 & 41.86 & 102.04 & 34.85 & 108.36 & 44.63 & 114.12 & 39.98 & 116.20 & 42.12 & 148.47\\
\hline
\end{tabular}}
\end{small}
\end{center}
\end{table*}

\begin{figure*}[t]
	\centering
	\includegraphics[width=0.98\textwidth]{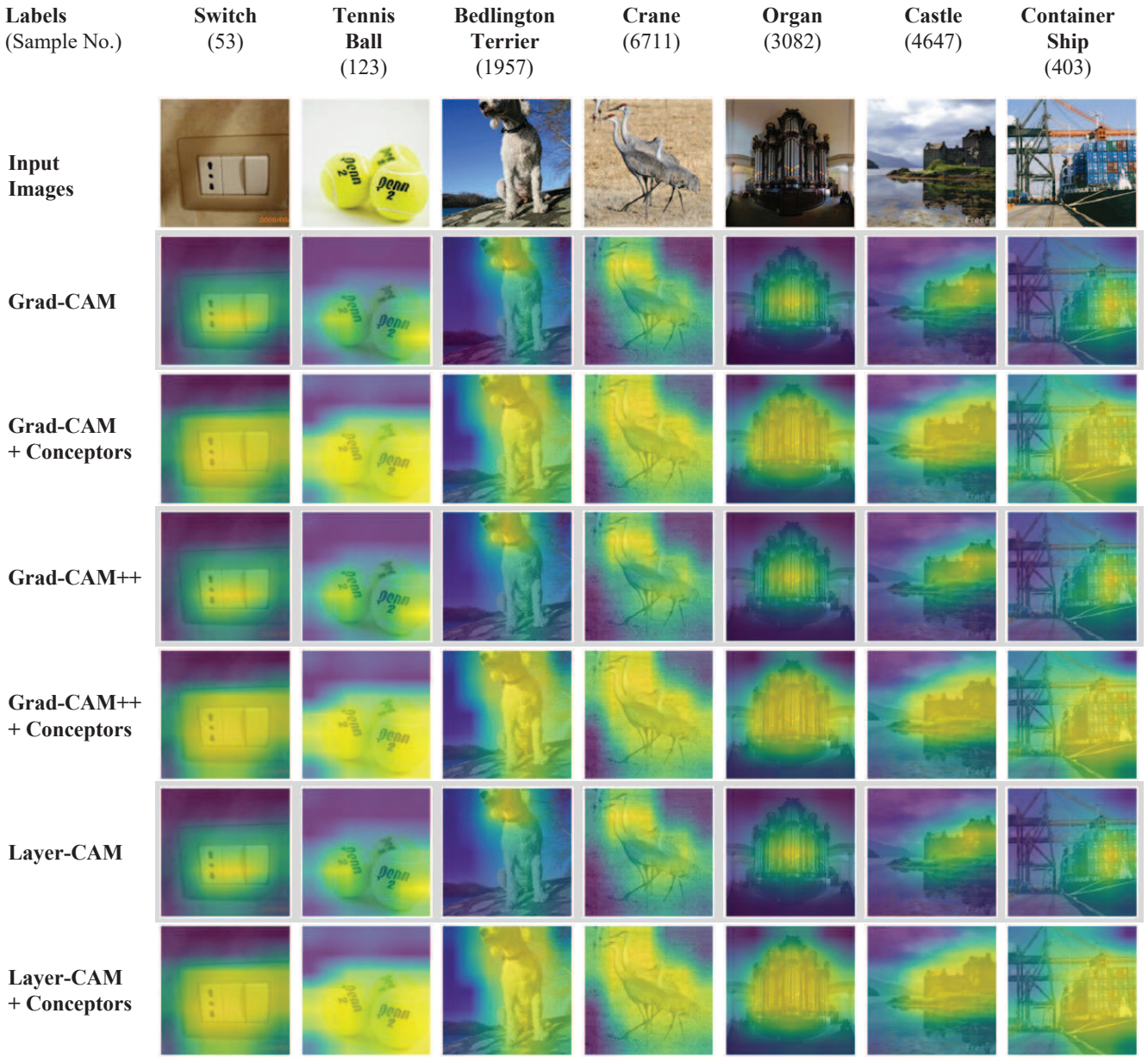}
	\caption{Examples of using Conceptors with conventional CAM-based Methods: In the conventional CAM-based methods, the focus has been put solely on the core parts (e.g., the centers of the tennis balls, the necks of the dog and crane), the inter-part collaborative relation has been ignored. With Conceptors, the inter-part relation has been recovered through the inter- and intra-channel relation modeling because parts are usually modeled by different channels or different regions inside a channel. Therefore, the target objects are better covered when the attentions on parts are connected through Conceptor learning.}\label{fig:CAMSample1}
\end{figure*}

\section{Experiments}
\label{sec:exp}
\subsection{Setup}
\label{sec:exp_setup}
To evaluate the performance of the proposed framework, we have conducted experiments on three popoular datasets of ImageNet Large-Scale  Visual  Recognition  Challenge  2012  (ILSVRC2012 validation set) \cite{russakovsky2015imagenet}, the PASCAL Visual Object Classes challenge (VOC2007 test set) \cite{Everingham10}, and Microsoft COCO (COCO2014 validation set) \cite{lin2014microsoft}.
Due to the space limitation and to be aligned to previous CAM studies, we will present and discuss the results mainly on ILSVRC2012. This is the default dataset if not mentioned specficially. The results on VOC and COCO will be given and disucssed when comparing to the state-of-the-art methods in Section~\ref{sec:exp_comparison_SOTA}.
For the experiments on ILSVRC2012, the full validation dataset of $50,000$ samples have been employed for comprehensiveness of the validation. Note that in some previous work, the results are reported on subsets of ILSVRC2012 (e.g., only $4\%$ samples used in \cite{wang2020score} for Score-CAM, \cite{wang2020ss} for SS-CAM and \cite{naidu2020cam} for IS-CAM respectively) and thus might be slightly different from those reported in this section.
All images have been reshaped to $224\times 224 \times 3$. Normalization has been done to regulate the values into the range of $[0,1]$ using the mean and standard deviation. All experiments have been performed on a machine with an Intel Xeon Gold 6248 CPU, a Nvidia Tesla T4 GPU, and 32G memory. The source code of the Conceptor-CAM can be found at https://www.github.com/(will make it open to public upon acceptance)

To measure the performance, we have employed two widely adopted metrics of Average Increase (AI) and Average Drop (AD) \cite{8354201,wang2020score,wang2020ss,desai_2020_WACV,naidu2020cam} as follows
\begin{equation}
\begin{split}
AI &= \left(\sum_{i=1}^N \frac{Sign(Y_i^c<\boldsymbol{\mathcal{S}}_{l \cdot i}^c)}{N}\right) \times 100, \\ 
AD &= \left(\sum_{i=1}^N \frac{max(0, Y_i^c-\boldsymbol{\mathcal{S}}_{l \cdot i}^c)}{Y_i^c}\right) \times 100, \\
\end{split}
\end{equation}

where $Sign(\cdot)$ denotes an indicator function that returns 1 when the expression inside (e.g., $Y_i^c<\boldsymbol{\mathcal{S}}_{l \cdot i}^c$) is true or 0 otherwise, $Y_i^c$ is the predicated score of class $c$ on $i$-th image in the test set of $N$ samples, and $\boldsymbol{\mathcal{S}}_{l \cdot i}^c$ is the predicated score of class $c$ on the corresponding $i$-th saliency map generated from the activation map $\boldsymbol{\mathcal{F}}_l^c$ at layer $l$. Note that AI is a positive index with which the larger the value is the better the performance is. We will use $\uparrow$ as the indicator for positive indices hereafter. Similarly, we use $\downarrow$ for negative indices such as the AD and the time cost.

\subsection{Conceptor-CAM}
\label{sec:exp_conceptor-cam}
\subsubsection{Compatibility with Existing CAM-based Methods}
\label{sec:exp_compatibility_cam-based}
In this section, we integrate the Conceptors into three popularly employed CAM-based methods, namely the Grad-CAM \cite{selvaraju2017grad}, Grad-CAM++ \cite{8354201}, and Score-CAM \cite{wang2020score}, to verify the compatibility. 
In addition, we have included Layer-CAM \cite{9462463}, SS-CAM \cite{wang2020ss} and Ablation-CAM \cite{desai_2020_WACV} as the representatives of the latest CAM-based methods.
The test has covered almost all well-recognized CAM methods except for the original CAM \cite{7780688}, because it requires the modification of the target networks by physically adding an average pooling layer. This requirement makes it less feasible to be implemented, and more importantly, it is unfair when comparing to other CAM-based methods because it has changed the inference path of the target networks.
Furthermore, we have normalized the feature maps using the hyperbolic tangent function (Tanh) to ease the calculation.
The results are shown in Table~\ref{tbl:cam-based_methods}. Note that we are using ResNet18 as the base model for this experiment because of its balance between efficiency and effectiveness. The comparison of base models will be given later in Section~\ref{sec:exp_compatibility_CNNs}.

It is easy to see that Concepotors are compatible with all CAM-based methods by gaining improvement ranging from $32.20\%$ to $116.20\%$ over runs without using Conceptors. This has confirmed the effectiveness of adding synchronization (through the inter- and intra-channel relations) to the CAM inference.

\begin{table*}[htbp]
\caption{Test of Compatibility of Conceptor Learning to popular CNN Models. The best results are in bold font.}
\label{tbl:different_CNNs}
\begin{center}
\begin{small}
\begin{tabular}{c|c|c|c|c|c|c|c|c|c|c|c|c}
\hline
& \multicolumn{3}{c|}{VGG16} & \multicolumn{3}{c|}{Inception-V3} & \multicolumn{3}{c|}{ResNet18} & \multicolumn{3}{c}{ResNet50} \\
\cline{2-13}
  Runs & \makecell[c]{AI $\uparrow$} & \makecell[c]{AD $\downarrow$} & \makecell[c]{Time $\downarrow$} & \makecell[c]{AI $\uparrow$} & \makecell[c]{AD $\downarrow$} & \makecell[c]{Time $\downarrow$} & \makecell[c]{AI $\uparrow$} & \makecell[c]{AD $\downarrow$} & \makecell[c]{Time $\downarrow$} & \makecell[c]{AI $\uparrow$} & \makecell[c]{AD $\downarrow$} & \makecell[c]{Time $\downarrow$}\\
\hline
\makecell[c]{without\\Conceptors} & 22.87 & 27.97 & 2.83 & 45.38 & 10.82 & 8.85 & 38.11 & 13.96 & 0.58 & 44.61 & 9.08 & 6.45\\
\hline
\makecell[c]{with Tanh+\\Conceptors} & 31.07 & 21.24 & 2.85 & \textbf{52.75} & 6.02 & 8.87 & 51.39 & 6.70 & \textbf{0.58} & 52.19 & \textbf{5.35} & 6.45\\
\hline
\makecell[c]{Improvement\\($\%$)} & 35.85 & 31.69 & -0.70 & 16.24 & 79.73 & -0.23 & 34.85 & 108.36 & 0 & 16.99 & 69.72 & 0\\
\hline
\end{tabular}
\end{small}
\end{center}
\end{table*}

This is more intuitively shown in Fig.~\ref{fig:CAMSample1}, where the performance of each method on various examples is given. Conventional CAM methods all tend to put the focus on only one part of the target objects while neglect the contribution of other parts (e.g., centers of the tennis balls in sample \#123, necks of the dog in sample \#1957 and the cranes in sample \#6711). As we mentioned in Section~\ref{sec:intra-channel}, this might be due to fact that the direct weighted fusion of channels will give attention to the large and salient parts while disregard these smaller and less dominating parts \cite{wei2017object,8237644}. The use of Conceptors has significantly addressed this issue in the sense that the target objects are better covered in the saliency maps. It is an indication that the attentions of the parts have been well synchronized through the inter- and intra-channel relations and thus these parts are connected as a whole in the saliency maps.

Comparing between CAM-based methods, in Table~\ref{tbl:cam-based_methods}, the Score-CAM has achieved the superior performance over the other two by $13.93\%$ ($31.95\%$) to $20.72\%$ ($42.34\%$) in Average Increase (Drop). Therefore, in the experiments thereafter, we will use Score-CAM as the baseline to study the factors related to Conceptor-CAM.
In addition, another fact we can learn from Table~\ref{tbl:cam-based_methods} is that the Tanh normalization is much more important for Grad-CAM and Grad-CAM++ when applying Conceptor-CAM. Otherwise, the performance gain is not guaranteed. This is not surprising, because it can be seen from Eq.~(\ref{eqn:cam-based}) that these two methods rely on the values of feature maps directly for calculation, which introduces larger variance than the difference of the feature maps that has been used by Score-CAM.

\begin{table*}[htbp]
\caption{Performance of using different evidences for Conceptor Learning. The best results are in bold font.}
\label{tbl:different_conceptors}
\begin{center}
\begin{small}
\begin{tabular}{c|c|c|c|c|c}
\hline
Model & Metric & \makecell[c]{Without \\ Conceptors} & \makecell[c]{Positive \\ Conceptors $\boldsymbol{{C}}$} & \makecell[c]{Complementary \\ Conceptors $\boldsymbol{{C}}^*$} & \makecell[c]{Comprehensive \\ Conceptors $\boldsymbol{{C}}+\boldsymbol{{C}}^*$} \\
\hline
ResNet18 & \makecell[c]{Average\\Increase $\uparrow$} & 38.11 & 51.39 & 51.50 & \textbf{54.55} \\
\cline{2-6}
& \makecell[c]{Average\\Drop $\downarrow$} &  13.96 & \textbf{6.70} & 9.51 &  7.41 \\
\hline
ResNet50 & \makecell[c]{Average\\Increase $\uparrow$} & 44.61 & 52.20 & 55.08 & \textbf{55.79} \\
\cline{2-6}
& \makecell[c]{Average\\Drop $\downarrow$} & 9.08 & 5.35 & 5.46 & \textbf{5.10} \\
\hline
\end{tabular}
\end{small}
\end{center}
\end{table*}

\subsubsection{Compatibility to CNN Models}
\label{sec:exp_compatibility_CNNs}
To verify the compatibility of Conceptors to existing CNN models and also to explain our selection of ResNet18 as the base model, we have applied Conceptors to VGG16 \cite{simonyan2014very}, Inception-V3 \cite{Szegedy_2016_CVPR}, ResNet18 \cite{he2016deep}, and ResNet50 \cite{he2016deep} respectively. The results are shown in Table~\ref{tbl:different_CNNs}.

It is not surprising that Conceptor-CAM has achieved a consistent improvement over the baseline runs across all types of CNN models by $35.85\%$ on VGG16, $16.24\%$ on Inception-V3, $34.85\%$ on ResNet18, and $16.99\%$ on ResNet50. It has validated that the compatibility and effectiveness of Conceptor-CAM are generalizable to prevalent CNN models. In addition, we can see that ResNet18 is the best balanced model between effectiveness and efficiency by demonstrating the acceptable performance in Average Increase ($65.40\%$ over VGG16) and the
best efficiency ($391.38\%$, $1429.31\%$, and $1012.07\%$ faster than VGG16, Inception-V3, ResNet50 respectively).

\subsubsection{Study of the Balancing Factor $\alpha$}
\label{sec:exp_alpha}
\begin{figure}
	\centering
	\includegraphics[width=1.02\textwidth]{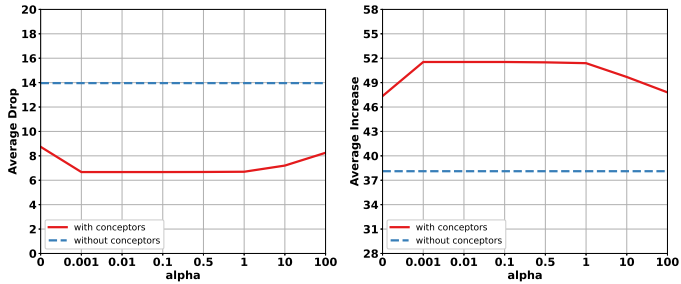}
	\caption{The performance of Conceptor-CAM over the balancing factor alpha: The performance is approaching optimal in the range of (0,1] but overall it is basically insensitive to the parameter alpha.}\label{fig:alpha}
\end{figure}

For more experience of using Conceptor-CAM, we study the only parameter $\alpha$ of the proposed framework in this section. The results of the performance over different settings of $\alpha$ in the range of $[0,100]$ is shown in Fig.~\ref{fig:alpha}.
Note that to get the solution of the Conceptor matrix using Eq.~(\ref{eqn:con}), we need to calculate the inverse of $(\boldsymbol{{R}}+\alpha^{-2}\boldsymbol{I})$. In most of the cases, this part is invertible, because the addition of  $\alpha^{-2}\boldsymbol{I}$ to $\boldsymbol{{R}}$ has significantly increased the probability that the result is a non-singular matrix. However, when $\alpha^{-2}$ is approaching zero, the result is more about $\boldsymbol{{R}}$ itself which is practically easy to be singular. We have to use the Moore-Penrose inverse in this case.

In Fig.~\ref{fig:alpha}, we have used different scales at the range of $(0,0.1)$ (when the regulation part dominating the synchronization part), $[0.1,1]$ (in case of $\alpha$ working as an ordinary balancing factor), and $(1,100]$ (for studying the case of $\alpha^{-2}$ approaching zero ).
We can see that the best performance can be obtained when $\alpha$ is in the range of $(0,1)$. It is consistent with our intuition that performance gain can only be obtained when the synchronization and regulation are well balanced.
It is slightly surprising that the result at the point $\alpha=0$ (we define $\alpha^{-2}=0$ when $\alpha=0$) is different from its theoretical equivalent (without Conceptors) by $38.11\%$ ($13.96\%$) in Average Increase (Drop). This is due to the Moore-Penrose inverse which makes the result of $\boldsymbol{{R}}(\boldsymbol{R})^{-1}$ not exactly the same as the theoretical solution (an identity matrix $\boldsymbol{I}$). Consequently, the synchronization, even might be modest, has been done which is beyond our theoretical expectation of doing nothing.
Besides, it is not surprising that the performance degrades gradually when the regulation is dominating the synchronization (i.e., $\alpha$ is in the range of $(0, 0.1)$. 
In the experiments, we have used the setting of $\alpha=1$ by default which will bring convenience for calculation and formal verification even it is not the best performance of Conceptor-CAM.

\subsubsection{Pseudo-negative Evidence}
\label{sec:exp_pseudo-negative}
To verify the effectiveness of using the pseudo-negative evidences and the advantage of Boolean supplement of the Conceptors, we have conducted experiments using the Conceptors ($\boldsymbol{{C}}$) (denote as the positive Conceptors to ease the description), the complementary Conceptors ($\boldsymbol{{C}}^*$), and the comprehensive Conceptors by fusing of the first two Conceptors ($\boldsymbol{{C}}+\boldsymbol{{C}}^*$) respectively. To be more extensive, we have also included the performance on ResNet50.
The results are shown in Table~\ref{tbl:different_conceptors}.

\begin{figure*}[t]
	\centering
	\includegraphics[width=0.80\textwidth]{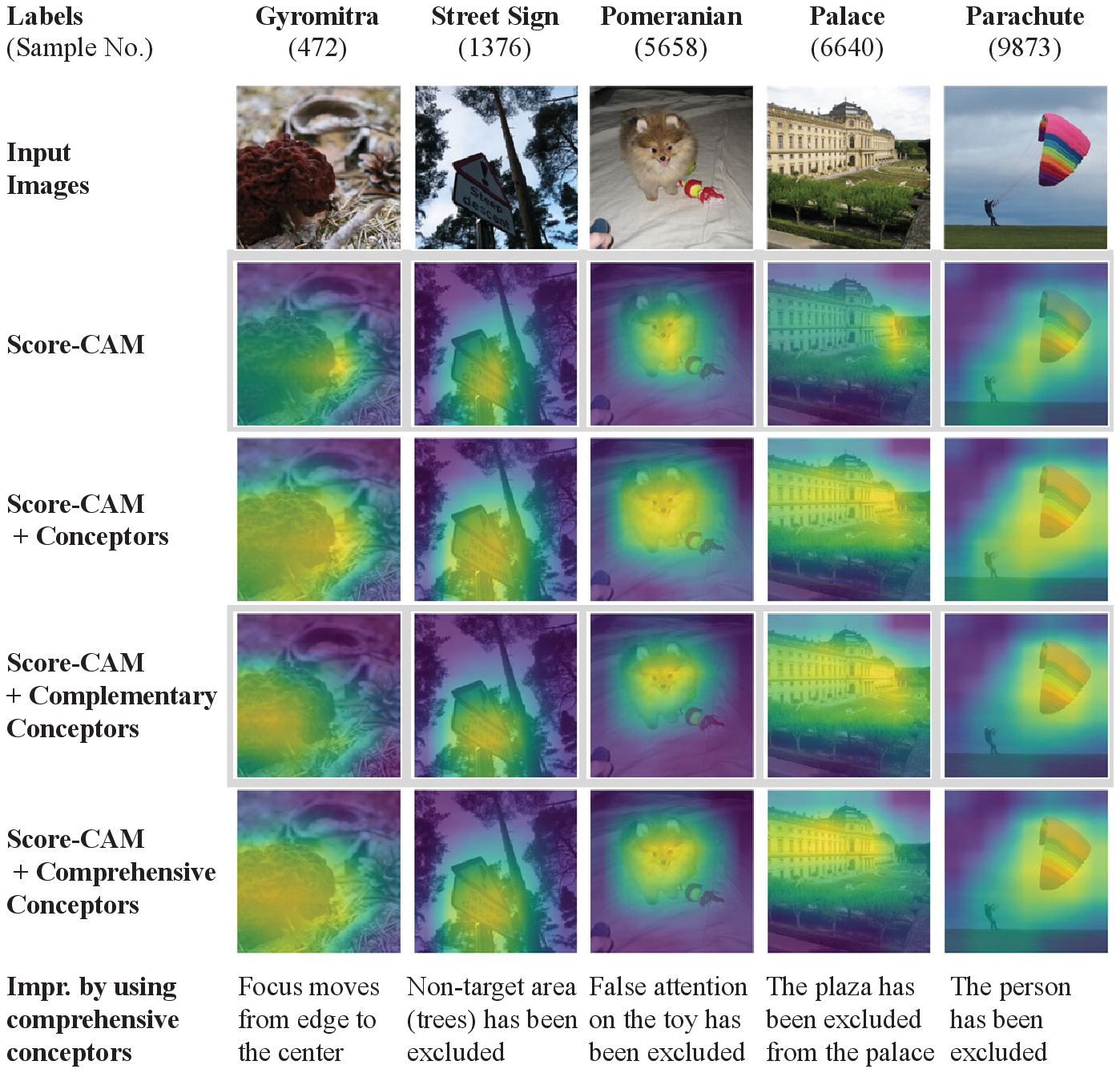}
	\caption{Examples of using different evidences for Conceptor Learning: The (positive) Conceptors tend to connect parts that have been ignored in conventional CAM-based methods (i.e., Score-CAM in this figure) and thus have obtained better coverage but unavoidably covered some non-target (background) areas. The complementary Conceptors learned through the pseudo-negative evidences are trying to exclude the non-target areas but sometimes will scarifies target parts along the boundaries. These two have been better balanced with the comprehensive Conceptors. }\label{fig:CAMSample2}
\end{figure*}

\begin{table*}[htbp]
\caption{Comparison of Conceptor-CAM to the state-of-the-art CAM-based methods on ILSVRC2012. The best results are in bold font.}
\label{tbl:SOTA}
\begin{center}
\begin{small}
\resizebox{\linewidth}{!}{
\begin{tabular}{c|c|c|c|c|c|c|c|c|c}
\hline
& \multicolumn{3}{c|}{Conceptor-CAM} & \multicolumn{3}{c|}{Popular CAMs} & \multicolumn{3}{c}{Latest CAMs}\\
\cline{2-10}
& \makecell[c]{Comprehensive\\$\boldsymbol{C}+\boldsymbol{C}^*$} & \makecell[c]{Complementary\\ $\boldsymbol{C}^*$} & \makecell[c]{Positive\\$\boldsymbol{C}$} & Grad-CAM & Grad-CAM++ & Score-CAM & Layer-CAM & SS-CAM & Ablation-CAM \\
\hline
\makecell[c]{Average\\Increase $\uparrow$} & \textbf{54.55} & 51.50 & 51.39 & 33.45 & 31.68 & 38.11 & 31.57 & 32.99 & 32.43 \\
\hline
\makecell[c]{Average\\Drop $\downarrow$} & 7.41 & 9.51 & \textbf{6.70} & 19.24 & 19.84 & 13.96 & 19.87 &  18.42 & 19.48\\
\hline
\end{tabular}}
\end{small}
\end{center}
\end{table*}

\begin{table*}[htbp]
\caption{Comparison of Conceptor-CAM to the state-of-the-art CAM-based methods on VOC. The best results are in bold font.}
\label{tbl:SOTA_VOC}
\begin{center}
\begin{small}
\resizebox{\linewidth}{!}{
\begin{tabular}{c|c|c|c|c|c|c|c|c|c}
\hline
& \multicolumn{3}{c|}{Conceptor-CAM} & \multicolumn{3}{c|}{Popular CAMs} & \multicolumn{3}{c}{Latest CAMs}\\
\cline{2-10}
& \makecell[c]{Comprehensive\\$\boldsymbol{C}+\boldsymbol{C}^*$} & \makecell[c]{Complementary\\ $\boldsymbol{C}^*$} & \makecell[c]{Positive\\$\boldsymbol{C}$} & Grad-CAM & Grad-CAM++ & Score-CAM & Layer-CAM & SS-CAM & Ablation-CAM \\
\hline
\makecell[c]{Average\\Increase $\uparrow$} & 51.37 & \textbf{53.23} & 46.51 & 42.43 & 37.34 & 46.12 & 38.07 & 44.10 & 39.76 \\
\hline
\makecell[c]{Average\\Drop $\downarrow$} & 3.65 & \textbf{3.44} & 4.90 & 13.68 & 10.88 & 5.06 & 10.70 & 5.92 & 16.24 \\
\hline
\end{tabular}}
\end{small}
\end{center}
\end{table*}

\begin{table*}[htbp]
\caption{Comparison of Conceptor-CAM to the state-of-the-art CAM-based methods on COCO. The best results are in bold font.}
\label{tbl:SOTA_COCO}
\begin{center}
\begin{small}
\resizebox{\linewidth}{!}{
\begin{tabular}{c|c|c|c|c|c|c|c|c|c}
\hline
& \multicolumn{3}{c|}{Conceptor-CAM} & \multicolumn{3}{c|}{Popular CAMs} & \multicolumn{3}{c}{Latest CAMs}\\
\cline{2-10}
& \makecell[c]{Comprehensive\\$\boldsymbol{C}+\boldsymbol{C}^*$} & \makecell[c]{Complementary\\ $\boldsymbol{C}^*$} & \makecell[c]{Positive\\$\boldsymbol{C}$} & Grad-CAM & Grad-CAM++ & Score-CAM & Layer-CAM & SS-CAM & Ablation-CAM \\
\hline
\makecell[c]{Average\\Increase $\uparrow$} & 64.31 & \textbf{65.49} & 59.86 & 52.72 & 51.19 & 55.77 & 49.87 & 54.08 & 52.23 \\
\hline
\makecell[c]{Average\\Drop $\downarrow$} & 5.99 & \textbf{5.89} & 6.33 & 17.79 & 13.23 & 8.69 & 13.59 & 9.27 & 18.05 \\
\hline
\end{tabular}}
\end{small}
\end{center}
\end{table*}

With the complementary Conceptors ($\boldsymbol{{C}}^*$), which has been generated with the pseudo-negative evidences using NOT operation defined on Conceptors, the performance is comparable to that using positive Conceptors ($\boldsymbol{{C}}$) with $6.15\%$ superiority in Average Increase but $10.60\%$ degration in Average Drop.
The performance obtained by fusing the positive and complementary Conceptors outperforms the others as expected (with $43.14\%$ ($88.39\%$) gain in Average Increase (Drop) over the run without Conceptors, and $5.92\%$ ($28.34\%$) over the complementary Conceptors). This has further confirmed the ability of pseudo-negative evidences to provide complementary information for the inference.

In Fig.~\ref{fig:CAMSample2}, we have compared the performance of these Conceptors on samples which will explain the rationale more intuitively.
Generally speaking, we can find from the examples that the positive Conceptors tend to connect the core parts to the other parts of the target objects and thus have a better coverage than that of the original Score-CAM which only focuses more on the core parts.
By contrast, the complementary Conceptors, which have been learned from the pseudo-negative evidences, tend to exclude the targets from the background. For example, on sample \#5658, the toy, which has been falsely covered by the original Score-CAM and the positive Conceptors, has been successfully excluded using the complementary Conceptors. Similar effect has been observed on sample \#9873 with exclusion of the person from the parachute, on sample \#6640 of the 
plaza from the palace, and on sample \#1376 of trees from the sign.
Furthermore, the complementary Conceptors have demonstrated a balanced effort of the positive and complementary Concepotrs to include and exclude the targets, and therefore obtained the best results. This is particularly obvious on samples \#472 where the original focus has been put on the edge of the gyromitra incorrectly, the complementary Conceptors have pushed the attention towards the central part but missed some parts of the target, and the comprehensive Conceptors have covered the gyromitra well and reduced the attention on the background part alone the edge (to which the positive Conceptors have given more attention). Similarly, on sample \#9873, the focus has been amended from the part between the parachute and the person to the parachute.

\subsection{Comparison to the State-of-the-art}
\label{sec:exp_comparison_SOTA}
We have compared the Conceptor-CAM to the State-of-art (SOTA) CAM-based methods including three popular methods of Grad-CAM \cite{8354201}, Grad-CAM++ \cite{selvaraju2017grad}, and Score-CAM \cite{wang2020score}, and three latest methods of SS-CAM \cite{wang2020ss}, Layer-CAM \cite{9462463} and Ablation-CAM \cite{desai_2020_WACV}. The results on ILSVRC2012 \cite{russakovsky2015imagenet}, VOC \cite{Everingham10}, and COCO \cite{lin2014microsoft} are shown in Table~\ref{tbl:SOTA}, Table~\ref{tbl:SOTA_VOC}, and Table~\ref{tbl:SOTA_COCO}, respectively.
We can see that the Conceptor-CAM methods outperform the SOTA methods by $43.14\%\sim 72.79\%$ ($88.39\%\sim 168.15\%$) on ILSVRC2012 in Average Increase (Drop), $15.42\%\sim 42.55\%$ ($47.09\%\sim 372.09\%$) on VOC, and $17.43\%\sim 31.32\%$ ($47.54\%\sim 206.45\%$) on COCO, respectively. The reason has already been discussed in previous sections.
However, the Complementary Conceptor-CAM works better than the other two on VOC and COCO this time. This is an observation different from those on ILSVRC2012. It is in fact not surprising, because the Complementary Conceptors are generated and improved from the positive Conceptors, which may have already collected enough clues in most of the cases.
Therefore, it could be considered as a more balanced solution.


\section{Conclusion}
In this paper, we have employed the Conceptor learning to model inter- and intra-channel relations for CAM inference. Both formal verification and experimental study have been provided to validate the effectiveness of the proposed Conceptor-CAM.
In addition, by using of pseudo-negative evidences, we have generated complementary Conceptors to provide more comprehensive CAM understanding of the DNN inference, which has confirmed the advantage of using Boolean operations in CAM learning.
Furthermore, we have proven that the Conceptor-CAM is an open framework that is compatible to conventional CAM-based methods and popular DNNs.

While encouraging results have been observed, the Conceptor-CAM has followed the conventional way of generating the saliency maps from the last convolutional block/layer of DNNs.
This might have ignored the inter-block/layer relation of DNN architectures. A recent study in \cite{9385115} has shown that aggregating CAM evidences from different stages (early or later layers along the feed-forward path) is able to generate more reliable results.
In the future, we will study whether this could be integrated into Conceptor-CAM in a formulated way.

\ifCLASSOPTIONcaptionsoff
  \newpage
\fi



\bibliographystyle{IEEEtran}
\bibliography{IEEEabrv,refs}

\begin{thebibliography}{10}
\providecommand{\url}[1]{#1}
\csname url@samestyle\endcsname
\providecommand{\newblock}{\relax}
\providecommand{\bibinfo}[2]{#2}
\providecommand{\BIBentrySTDinterwordspacing}{\spaceskip=0pt\relax}
\providecommand{\BIBentryALTinterwordstretchfactor}{4}
\providecommand{\BIBentryALTinterwordspacing}{\spaceskip=\fontdimen2\font plus
\BIBentryALTinterwordstretchfactor\fontdimen3\font minus
  \fontdimen4\font\relax}
\providecommand{\BIBforeignlanguage}[2]{{%
\expandafter\ifx\csname l@#1\endcsname\relax
\typeout{** WARNING: IEEEtran.bst: No hyphenation pattern has been}%
\typeout{** loaded for the language `#1'. Using the pattern for}%
\typeout{** the default language instead.}%
\else
\language=\csname l@#1\endcsname
\fi
#2}}
\providecommand{\BIBdecl}{\relax}
\BIBdecl

\bibitem{7780688}
B.~Zhou, A.~Khosla, A.~Lapedriza, A.~Oliva, and A.~Torralba, ``Learning deep
  features for discriminative localization,'' in \emph{2016 IEEE Conference on
  Computer Vision and Pattern Recognition (CVPR)}, 2016, pp. 2921--2929.

\bibitem{selvaraju2017grad}
R.~R. Selvaraju, M.~Cogswell, A.~Das, R.~Vedantam, D.~Parikh, and D.~Batra,
  ``Grad-cam: Visual explanations from deep networks via gradient-based
  localization,'' in \emph{Proceedings of the IEEE international conference on
  computer vision}, 2017, pp. 618--626.

\bibitem{wang2020score}
H.~Wang, Z.~Wang, M.~Du, F.~Yang, Z.~Zhang, S.~Ding, P.~Mardziel, and X.~Hu,
  ``Score-cam: Score-weighted visual explanations for convolutional neural
  networks,'' in \emph{Proceedings of the IEEE/CVF Conference on Computer
  Vision and Pattern Recognition Workshops}, 2020, pp. 24--25.

\bibitem{zeiler2014visualizing}
M.~D. Zeiler and R.~Fergus, ``Visualizing and understanding convolutional
  networks,'' in \emph{European conference on computer vision}.\hskip 1em plus
  0.5em minus 0.4em\relax Springer, 2014, pp. 818--833.

\bibitem{wei2017object}
Y.~Wei, J.~Feng, X.~Liang, M.-M. Cheng, Y.~Zhao, and S.~Yan, ``Object region
  mining with adversarial erasing: A simple classification to semantic
  segmentation approach,'' in \emph{Proceedings of the IEEE conference on
  computer vision and pattern recognition}, 2017, pp. 1568--1576.

\bibitem{8237644}
D.~Kim, D.~Cho, D.~Yoo, and I.~S. Kweon, ``Two-phase learning for weakly
  supervised object localization,'' in \emph{2017 IEEE International Conference
  on Computer Vision (ICCV)}, 2017, pp. 3554--3563.

\bibitem{li2018tell}
K.~Li, Z.~Wu, K.-C. Peng, J.~Ernst, and Y.~Fu, ``Tell me where to look: Guided
  attention inference network,'' in \emph{Proceedings of the IEEE Conference on
  Computer Vision and Pattern Recognition}, 2018, pp. 9215--9223.

\bibitem{jaeger2014controlling}
H.~Jaeger, ``Controlling recurrent neural networks by conceptors,'' \emph{arXiv
  preprint arXiv:1403.3369}, 2014.

\bibitem{jaeger2017using}
H.~Jaeger, ``Using conceptors to manage neural long-term memories for temporal
  patterns,'' \emph{The Journal of Machine Learning Research}, vol.~18, no.~1,
  pp. 387--429, 2017.

\bibitem{8354201}
A.~Chattopadhay, A.~Sarkar, P.~Howlader, and V.~N. Balasubramanian,
  ``Grad-cam++: Generalized gradient-based visual explanations for deep
  convolutional networks,'' in \emph{2018 IEEE Winter Conference on
  Applications of Computer Vision (WACV)}, 2018, pp. 839--847.

\bibitem{wang2020ss}
H.~Wang, R.~Naidu, J.~Michael, and S.~S. Kundu, ``Ss-cam: Smoothed score-cam
  for sharper visual feature localization,'' \emph{arXiv preprint
  arXiv:2006.14255}, 2020.

\bibitem{9462463}
P.-T. Jiang, C.-B. Zhang, Q.~Hou, M.-M. Cheng, and Y.~Wei, ``Layercam:
  Exploring hierarchical class activation maps for localization,'' \emph{IEEE
  Transactions on Image Processing}, vol.~30, pp. 5875--5888, 2021.

\bibitem{desai_2020_WACV}
S.~Desai and H.~G. Ramaswamy, ``Ablation-cam: Visual explanations for deep
  convolutional network via gradient-free localization,'' in \emph{Proceedings
  of the IEEE/CVF Winter Conference on Applications of Computer Vision (WACV)},
  March 2020.

\bibitem{9385115}
X.-Y. Wei, Z.-Q. Yang, X.-L. Zhang, G.~Liao, A.-L. Sheng, S.~K. Zhou, Y.~Wu,
  and L.~Du, ``Deep collocative learning for immunofixation electrophoresis
  image analysis,'' \emph{IEEE Transactions on Medical Imaging}, vol.~40,
  no.~7, pp. 1898--1910, 2021.

\bibitem{9069411}
G.~Cheng, J.~Yang, D.~Gao, L.~Guo, and J.~Han, ``High-quality proposals for
  weakly supervised object detection,'' \emph{IEEE Transactions on Image
  Processing}, vol.~29, pp. 5794--5804, 2020.

\bibitem{8847950}
H.-T. Joo and K.-J. Kim, ``Visualization of deep reinforcement learning using
  grad-cam: How ai plays atari games?'' in \emph{2019 IEEE Conference on Games
  (CoG)}, 2019, pp. 1--2.

\bibitem{8821313}
S.-H. Gao, M.-M. Cheng, K.~Zhao, X.-Y. Zhang, M.-H. Yang, and P.~Torr,
  ``Res2net: A new multi-scale backbone architecture,'' \emph{IEEE Transactions
  on Pattern Analysis and Machine Intelligence}, vol.~43, no.~2, pp. 652--662,
  2021.

\bibitem{pmlr-v97-tan19a}
M.~Tan and Q.~Le, ``{E}fficient{N}et: Rethinking model scaling for
  convolutional neural networks,'' in \emph{Proceedings of the 36th
  International Conference on Machine Learning}, ser. Proceedings of Machine
  Learning Research, K.~Chaudhuri and R.~Salakhutdinov, Eds., vol.~97.\hskip
  1em plus 0.5em minus 0.4em\relax PMLR, 09--15 Jun 2019, pp. 6105--6114.

\bibitem{ozturk2020automated}
T.~Ozturk, M.~Talo, E.~A. Yildirim, U.~B. Baloglu, O.~Yildirim, and U.~R.
  Acharya, ``Automated detection of covid-19 cases using deep neural networks
  with x-ray images,'' \emph{Computers in biology and medicine}, vol. 121, p.
  103792, 2020.

\bibitem{GARGEYA2017962}
R.~Gargeya and T.~Leng, ``Automated identification of diabetic retinopathy
  using deep learning,'' \emph{Ophthalmology}, vol. 124, no.~7, pp. 962--969,
  2017.

\bibitem{WANG2016237}
L.~Wang, Z.~Wang, and S.~Liu, ``An effective multivariate time series
  classification approach using echo state network and adaptive differential
  evolution algorithm,'' \emph{Expert Systems with Applications}, vol.~43, pp.
  237--249, 2016.

\bibitem{QIAN20181034}
G.~Qian and L.~Zhang, ``A simple feedforward convolutional conceptor neural
  network for classification,'' \emph{Applied Soft Computing}, vol.~70, pp.
  1034--1041, 2018.

\bibitem{10.1007/978-3-319-59081-3_35}
G.~Qian, L.~Zhang, and Q.~Zhang, ``Fast conceptor classifier in pre-trained
  neural networks for visual recognition,'' in \emph{Advances in Neural
  Networks - ISNN 2017}, F.~Cong, A.~Leung, and Q.~Wei, Eds.\hskip 1em plus
  0.5em minus 0.4em\relax Cham: Springer International Publishing, 2017, pp.
  290--298.

\bibitem{zhang2020chaotic}
A.~Zhang and Z.~Xu, ``Chaotic time series prediction using phase space
  reconstruction based conceptor network,'' \emph{Cognitive Neurodynamics},
  vol.~14, no.~6, pp. 849--857, 2020.

\bibitem{8894427}
Z.~Xu, L.~Zhong, and A.~Zhang, ``Phase space reconstruction-based conceptor
  network for time series prediction,'' \emph{IEEE Access}, vol.~7, pp.
  163\,172--163\,179, 2019.

\bibitem{he2018overcoming}
X.~He and H.~Jaeger, ``Overcoming catastrophic interference using
  conceptor-aided backpropagation,'' in \emph{International Conference on
  Learning Representations}, 2018.

\bibitem{7875131}
M.~Chen, M.~Mozaffari, W.~Saad, C.~Yin, M.~Debbah, and C.~S. Hong, ``Caching in
  the sky: Proactive deployment of cache-enabled unmanned aerial vehicles for
  optimized quality-of-experience,'' \emph{IEEE Journal on Selected Areas in
  Communications}, vol.~35, no.~5, pp. 1046--1061, 2017.

\bibitem{Liu_Ungar_Sedoc_2019}
T.~Liu, L.~Ungar, and J.~Sedoc, ``Unsupervised post-processing of word vectors
  via conceptor negation,'' \emph{Proceedings of the AAAI Conference on
  Artificial Intelligence}, vol.~33, no.~01, pp. 6778--6785, Jul. 2019.

\bibitem{qian2019single}
G.~Qian, L.~Zhang, and Y.~Wang, ``Single-label and multi-label conceptor
  classifiers in pre-trained neural networks,'' \emph{Neural Computing and
  Applications}, vol.~31, no.~10, pp. 6179--6188, 2019.

\bibitem{russakovsky2015imagenet}
O.~Russakovsky, J.~Deng, H.~Su, J.~Krause, S.~Satheesh, S.~Ma, Z.~Huang,
  A.~Karpathy, A.~Khosla, M.~Bernstein \emph{et~al.}, ``Imagenet large scale
  visual recognition challenge,'' \emph{International Journal of Computer
  Vision}, vol. 115, no.~3, pp. 211--252, 2015.

\bibitem{Everingham10}
M.~Everingham, L.~Van~Gool, C.~K.~I. Williams, J.~Winn, and A.~Zisserman, ``The
  pascal visual object classes (voc) challenge,'' \emph{International Journal
  of Computer Vision}, vol.~88, no.~2, pp. 303--338, Jun. 2010.

\bibitem{lin2014microsoft}
\BIBentryALTinterwordspacing
T.-Y. Lin, M.~Maire, S.~Belongie, J.~Hays, P.~Perona, D.~Ramanan, P.~Dollar,
  and L.~Zitnick, ``Microsoft coco: Common objects in context,'' in
  \emph{ECCV}.\hskip 1em plus 0.5em minus 0.4em\relax European Conference on
  Computer Vision, September 2014. [Online]. Available:
  \url{https://www.microsoft.com/en-us/research/publication/microsoft-coco-common-objects-in-context/}
\BIBentrySTDinterwordspacing

\bibitem{naidu2020cam}
R.~Naidu, A.~Ghosh, Y.~Maurya, S.~S. Kundu \emph{et~al.}, ``Is-cam: Integrated
  score-cam for axiomatic-based explanations,'' \emph{arXiv preprint
  arXiv:2010.03023}, 2020.

\bibitem{simonyan2014very}
K.~Simonyan and A.~Zisserman, ``Very deep convolutional networks for
  large-scale image recognition,'' \emph{arXiv preprint arXiv:1409.1556}, 2014.

\bibitem{Szegedy_2016_CVPR}
C.~Szegedy, V.~Vanhoucke, S.~Ioffe, J.~Shlens, and Z.~Wojna, ``Rethinking the
  inception architecture for computer vision,'' in \emph{Proceedings of the
  IEEE Conference on Computer Vision and Pattern Recognition (CVPR)}, June
  2016.

\bibitem{he2016deep}
K.~He, X.~Zhang, S.~Ren, and J.~Sun, ``Deep residual learning for image
  recognition,'' in \emph{Proceedings of the IEEE conference on computer vision
  and pattern recognition}, 2016, pp. 770--778.

\end{thebibliography}
\end{document}